\documentclass{article}

      \usepackage[final]{neurips_2024}

\usepackage[utf8]{inputenc} 
\usepackage[T1]{fontenc}    
\usepackage{hyperref}       
\usepackage{url}            
\usepackage{booktabs}       
\usepackage{amsfonts}       
\usepackage{graphicx}       
\usepackage{nicefrac}       
\usepackage{microtype}      
\usepackage{xcolor}         
\usepackage{amsmath} 
\usepackage{subcaption}

\title{Generating Auxiliary Tasks with Reinforcement Learning
}

\author{%
  Judah Goldfeder$^*$ \\
  Department of Computer Science \\
  Columbia University \\
  \texttt{jag2396@columbia.edu}
  \And
  Matthew So$^*$ \\
  Department of Computer Science \\
  Columbia University \\
  \texttt{ms5513@columbia.edu}
  \And
  Hod Lipson \\
  Department of Mechanical Engineering \\
  Columbia University \\
  \texttt{hod.lipson@columbia.edu}
}

\begin{document}

\def\thefootnote{*}\footnotetext{These authors contributed equally to this work}\def\thefootnote{\arabic{footnote}}
\maketitle

\begin{abstract}
Auxiliary Learning (AL) is a form of multi-task learning in which a model trains on auxiliary tasks to boost performance on a primary objective. While AL has improved generalization across domains such as navigation, image classification, and NLP, it often depends on human-labeled auxiliary tasks that are costly to design and require domain expertise. Meta-learning approaches mitigate this by learning to generate auxiliary tasks, but typically rely on gradient based bi-level optimization, adding substantial computational and implementation overhead. We propose RL-AUX, a reinforcement-learning (RL) framework that dynamically creates auxiliary tasks by assigning auxiliary labels to each training example, rewarding the agent whenever its selections improve the performance on the primary task. We also explore learning per-example weights for the auxiliary loss. On CIFAR-100 grouped into 20 superclasses, our RL method outperforms human-labeled auxiliary tasks and matches the performance of a prominent bi-level optimization baseline. We present similarly strong results on other classification datasets. These results suggest RL is a viable path to generating effective auxiliary tasks.
\end{abstract}

\section{Introduction}

Auxiliary Learning (AL) is a technique by which learned or pre-labeled auxiliary tasks are provided as an additional objective to a network during its training with the intended goal of improving the network's performance on a desired primary task. Auxiliary Learning can be thought of as a sub-field of multi-task learning, in which the objective of the training is to improve the main network's performance on the primary task while the auxiliary tasks regularize the training \citep{Caruana1997, DBLP:journals/corr/multitask}. It has been demonstrated that the inclusion of auxiliary tasks during training improves generalization and network performance on unseen samples across a large range of domains, including speech recognition, navigation, and image classification \citep{DBLP:journals/corr/JaderbergMCSLSK16, DBLP:journals/corr/goyal, DBLP:journals/corr/MirowskiPVSBBDG16, DBLP:journals/corr/abs-1805-06334-liebel, DBLP:journals/corr/ToshniwalTLL17}. Even small, tangentially related tasks have been shown to provide significant support to the main task \citep{DBLP:journals/corr/abs-1805-06334-liebel}. The intuition is that using the auxiliary task pushes the network to learn a shared representation of the data that guards against overfitting on the primary task \citep{DBLP:journals/corr/MAXL}.

A historic weakness of Auxiliary Learning has been the need for additional human labeling during the creation of supervised auxiliary tasks. This requires a large amount of human effort and domain expertise for each auxiliary task. Furthermore, in the multi-task context, we know that poor task selection can ultimately harm primary task performance \citep{DBLP:journals/corr/Gururangan}. Therefore, the manner in which primary and auxiliary tasks should be optimally combined during the weight update procedure can be ambiguous and require expert knowledge.

Meta Auxiliary Learning (MAXL) attempts to alleviate the problem of auxiliary task labeling through the procedural generation of an auxiliary task that optimizes the performance on the given primary task \citep{DBLP:journals/corr/MAXL}. The MAXL framework works by organizing the inputs of the primary task into hierarchical subclasses for each primary class using an additional label generation network. As such, MAXL is one of several approaches for dynamic label generation that run into the Bi-Level Optimization problem \citep{DBLP:journals/corr/auxilearn, Joint_data_task}.  Bi-Level Optimization, which is at the heart of many Meta Learning procedures, arises here as the gradients of the label network are calculated with respect to the performance of the main network on the primary task, resulting in a Hessian-inverse vector calculation and increased implementation complexity. 

In this work, we will attempt to bypass the need for this complex Bi-Level Optimization altogether by training a Reinforcement Learning (RL) agent to learn the auxiliary task selection (RL-AUX). Our approach involves encapsulating the main network's training loop (for both the primary and auxiliary tasks) within a Reinforcement Learning environment. We then expose the input data points as the environment's state and the network's performance on the main task as the reward. The environment's action space will enable the RL agent to provide auxiliary labels per data point. An agent trained in this environment should therefore learn the auxiliary task labeling that maximizes the performance of the main network on the primary task. We focus our work on commonly used image classification datasets, primarily CIFAR-100. Our experiment results reveal that the RL approach performs as well as the state-of-the-art MAXL approach with less model customization for the Hessian calculation.

After this initial success with RL labeling of the auxiliary task, we extend the work to explore loss weight assignment per data point. Several techniques have been explored in the literature to optimize auxiliary task weighting but none attempt to dynamically select the auxiliary task labels and also the weights at a sample-level \citep{kung-etal-2021-efficient, grégoire2023samplelevelweightingmultitasklearning, abbas, DBLP:Chennupati}.

To this end, this paper also presents a weight-aware version of MAXL (WA-MAXL) and a weight-aware version of our Reinforcement Learning architecture (WA-RLAUX). Both are shown to have significant improvement over their statically weighted counterparts while not requiring any additional hyperparameters or inducing a significant scaling cost.

In summary, this paper accomplishes two things that have not yet been presented in the literature to our knowledge. Firstly, we show that Reinforcement Learning can be used for auxiliary task selection while retaining the performance gains of the main network on the primary task. Moreover, it can do this while providing linear scaling with respect to the number of parameters in the main network. Secondly, we demonstrate methods to dynamically learn sample-level loss weights and an auxiliary task at the same time in both the Bi-Level Optimization and Reinforcement Learning approaches. We see significant performance improvements in both domains.

\section{Related Work}

\begin{figure}[htbp]
    \centering
    
    \begin{subfigure}[b]{0.48\linewidth} 
        \centering
        \includegraphics[width=\linewidth]{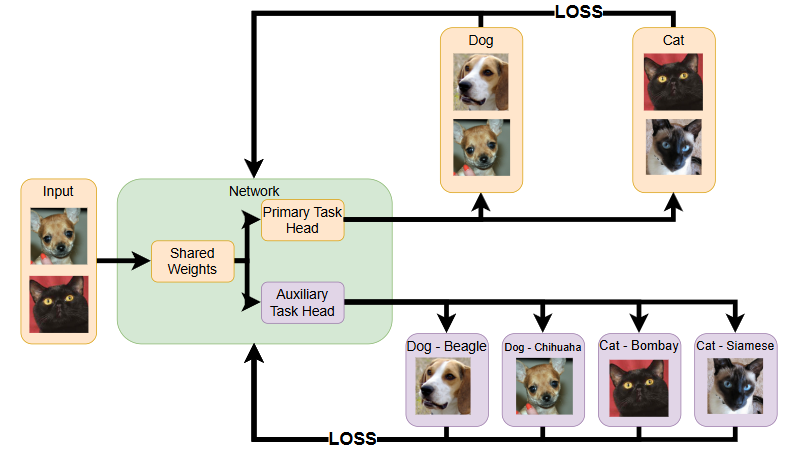}
        \caption{Classic Auxiliary Task Setup}
        \label{fig:ClassicAuxiliaryTask}
    \end{subfigure}
    \hfill                   
    \begin{subfigure}[b]{0.48\linewidth}
        \centering
        \includegraphics[width=\linewidth]{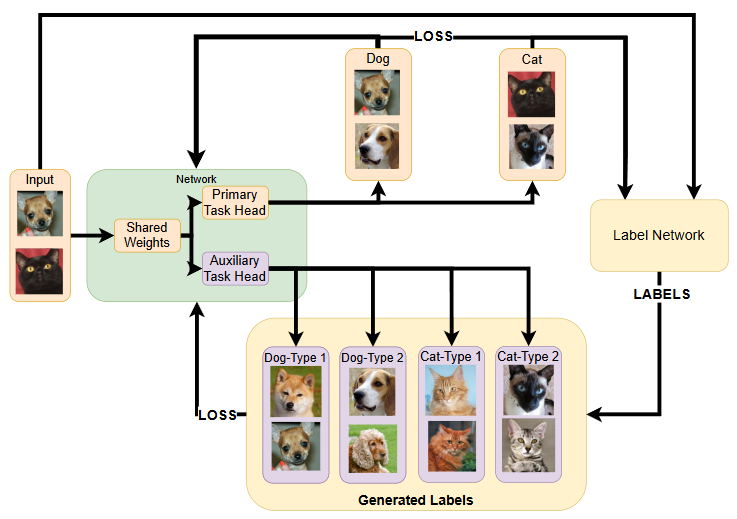}  
        \caption{Label Network Auxiliary Task Setup}
        \label{fig:LabelNetworkAuxiliaryTask}
    \end{subfigure}

    \caption{Sample Auxiliary Task Approaches}
    \label{fig:AuxTaskComparison}
\end{figure}

\textbf{Auxiliary and Multi-Task Learning in Vision}
Multi-task Learning (MTL) is a well-studied and widely-used Machine Learning method to have a network learn multiple tasks simultaneously \citep{Caruana1997, DBLP:journals/corr/abs-1805-06334-liebel, DBLP:journals/corr/ZhangY17aa}. Auxiliary Learning (AL) is a special case of MTL, in which there is one primary task of importance and one or more auxiliary tasks that support the performance of the main task \citep{DBLP:journals/corr/MAXL}. MTL and AL has been shown to improve target task performance compared to networks that are trained on a single task, particularly in low data contexts \citep{DBLP:journals/corr/standley, DBLP:journals/corr/ZhangY17aa}. There has been extensive work using Multi-task and Auxiliary Learning in the vision domain, such as the auxiliary classifiers in GoogleLeNet \citep{DBLP:journals/corr/GoogleLeNet}, multi-task cascaded convolutional networks \citep{DBLP:journals/corr/ZhangZL016}, state-of-the-art performance on three vision tasks using one convolutional network \citep{DBLP:journals/corr/EigenF14}, and many other examples \citep{MATTE, ubernet, DBLP:journals/corr/ladder}. Figure \ref{fig:ClassicAuxiliaryTask} represents a common, simple example of an auxiliary task in a classification task setting where the auxiliary task is provided by human labeling.

\textbf{Task Weighting}
The standard loss formulation in Auxiliary Learning is given by:
\[
\mathcal{L}_{total}(\theta) = \mathcal{L}_{primary}(\theta) + \sum_{i=1}^k \lambda_i \mathcal{L}_{aux}^{(i)}(\theta)
\]

Where $k$ is the number of auxiliary tasks, $\lambda_i$ represents the weight of the task $i$, and the network is parametrized by $\theta$. The Taskonomy framework provides a methodology to investigate the relationship between candidate auxiliary tasks and map their relationship based on their contribution to target task performance \citep{DBLP:journals/corr/taskonomy}. \citep{DBLP:journals/corr/standley} took this a step further and demonstrated that not all possible tasks are helpful for learning. As such, several papers in the field have attempted to learn optimal $\lambda_i$ loss weights on a per-task basis . \citep{DBLP:journals/corr/KendallGC17} introduced the idea of using homoscedastic aleatoric uncertainty to weight known auxiliary tasks. The auxiliary loss in their framework is given by:

\[\mathcal{L}(\theta)=\frac{1}{2 \sigma^2_{primary}} \mathcal{L}_{primary}(\theta) + log(\sigma_{primary}) + \sum_{i=1}^k[\frac{1}{2 \sigma^2_{i}} \mathcal{L}_{aux}(\theta) + log(\sigma_{aux}^{(i)})]  \]

Where the $\sigma$ values represent task-specific uncertainty noise. \citep{DBLP:journals/corr/abs-1805-06334-liebel} extended this work but enforced positive regularization values to achieve improved results \citep{Comparison}.
GradNorm demonstrated strong results by normalizing gradient magnitudes on a per-task basis \citep{DBLP:journals/corr/grad_norm} . Task weighting as a Pareto multi-objective optimization was also attempted \citep{DBLP:journals/corr/pareto}. 

While the previously mentioned approaches attempt to weight on a per-task basis, SLGrad presents a sample-level weighting approach for known auxiliary tasks that scales each sample on the cosine-similiarity of the sample's loss gradient and the primary task's validation gradient \citep{grégoire2023samplelevelweightingmultitasklearning}. Auxilearn also presents a framework to learn sample-specific weights for a known task using Bi-Level Optimization with implicit differentiation \citep{DBLP:journals/corr/auxilearn}.

\textbf{Task Generation}
MAXL started the label network paradigm for dynamic task generation in the Auxiliary Learning space using Bi-Level Optimization \citep{DBLP:journals/corr/MAXL}. The MAXL framework trains a label network to create an auxiliary task that optimizes the main network's performance on the target primary task. Figure \ref{fig:LabelNetworkAuxiliaryTask} shows how the setup of a label network is used to generate the auxiliary task. Bi-Level Optimization approaches attempt to find the optimal primary network weights $\theta^*$ and auxiliary labeling network weights $\phi^*$ that satisfy:
\[ \phi^*=\arg \min_{\phi} \mathcal{L}_{aux}(\theta^*(\phi)) \text{ s.t. } \theta^*(\phi) = \arg \min_{\theta} \mathcal{L}_{primary}(\theta,\phi)\]

The gradient update of the label network is given by:

\[ \nabla_{\phi} \mathcal{L}_{aux} (\theta^*(\phi)) = - \nabla_{\theta} \mathcal{L}_{aux} \cdot (\nabla^2_{\theta} \mathcal{L}_{primary})^{-1} \cdot \nabla_{\phi} \nabla_{\theta} \mathcal{L}_{primary} \]

Auxilearn, another approach using Bi-Level Optimization, attempted to use Neumann approximation to optimize this complex update \citep{DBLP:journals/corr/auxilearn}. Other Bi-Level Optimization approaches in task creation involve generating features/samples on the fly and finding useful "questions" as general value functions \citep{DBLP:journals/corr/abs-1909-04607, Joint_data_task}.

There have been several attempts at task generation without Bi-Level Optimization. These approaches involve selecting task objectives from a predefined or procedurally generated pool of tasks. Approaches include using Beta-Bernoulli multi-armed bandit framing \citep{DBLP:journals/corr/abs-1904-04153}, search over unified taxonomy \citep{dery2023aang}, and trial-and-error search over generated features \citep{pmlr-v232-rafiee23a}.

\textbf{Combining Auxiliary Learning and Reinforcement Learning}
The majority of work combining Reinforcement Learning with Auxiliary Learning involves constructing auxiliary tasks for RL agents to perform better on their assigned task \citep{DBLP:journals/corr/JaderbergMCSLSK16, pmlr-v80-riedmiller18a, ijcai2017p353}. 

There has been limited work in using Reinforcement Learning to improve the performance of networks through Multi-task and Auxiliary Learning. AutoAugment attempts to use RL to find optimal data augmentation strategies during training to improve a network's classification performance \citep{DBLP:journals/corr/abs-1805-09501}. In \citep{fan2018learning}, the researchers experiment using an RL teacher to select which data points from the training set should be used to train a main network.

\section{Problem Formulation}

This work focuses on using Reinforcement Learning to design auxiliary tasks for image classification. Let our dataset be represented as $D =\{(x_i,y_i)\}_{i=1}^N, x \in X, y \in Y$. We construct a policy $\pi_{\phi} : S \to A$, parameterized by $\phi$, that maps a state $s\in S$ to an action $a \to A$. In practice, the state is an input value of the dataset, i.e. $S=X$, and the action space defines another classification problem. We will train a main network $f$, parameterized by $\theta$, that has two output heads $f^{\theta}_{primary}:X \to Y$ and  $f^{\theta}_{aux}:X \to A$. We will train the network with the following loss:

\begin{equation}
    \mathcal{L}_{total}(\theta) = \mathcal{L}_{primary}(\theta) + \lambda \mathcal{L}_{aux}(\theta)
    \label{eq:rl_loss}
\end{equation}

Where $\lambda$ is the weight of the auxiliary task. We can pose the total loss per-sample as:
\begin{equation}
\ell_{total}(x_i, y_i) = \ell_{primary}(f_{primary}^{\theta}(x_i), y_i) + \lambda \ell_{aux}(f_{aux}^{\theta}(x_i),\pi_{\phi}(x_i))
    \label{eq:sample_rl_loss}
\end{equation}

We define $l_{primary}$ and $l_{aux}$ separately, as the primary and auxiliary task do not need to use the same metric in our framework.

\begin{figure}
    \centering
    \includegraphics[width=0.88\linewidth]{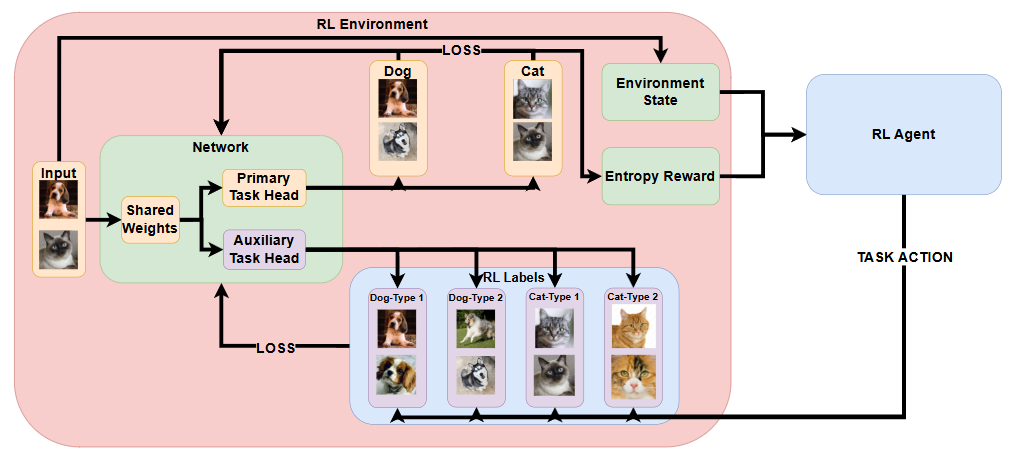}
    \caption{Reinforcement Learning Auxiliary Task Setup}
    \label{fig:rl_auxiliary_task}
\end{figure}

\textbf{Reinforcement Learning Approach}
The policy $\pi_{\phi}$ will be iteratively updated through an RL approach. To do this, we encapsulate the learning of the main network within an environment. Figure \ref{fig:rl_auxiliary_task} provides an overview of how the training is set up.

The policy $\pi_{\phi}$  provides a single auxiliary label for one data point in each step. A reward is provided to the policy after it labels $B_T$ samples, which is the training batch size for the main network. To calculate the reward, the environment trains the main network on the batch that the policy has labeled. The environment trains the network on both the primary and auxiliary tasks using the loss given by Equation \ref{eq:sample_rl_loss}. Then, a randomized evaluation batch of size $B_R$ is selected from the training sample. The reward provided to the policy will be the negative loss of the main network on the primary task over the evaluation batch with an added batch-wise entropy loss factor. The entropy loss factor $\mathcal{H}(\pi_\theta(x)_{(b)})$ per evaluation batch $b$ is given by:

\begin{equation}
\mathcal{H}(\pi_\theta(x)_{(b)}) = \sum_{k=1}^K \pi_\theta(x)_{(b)}^k \log \pi_\theta(x)_{(b)}^k, \pi_\theta(x)_{(b)}^k= \frac{1}{B_R} \sum_{n=1}^{B_R} \pi_\theta(x)_{(b)}^k[n] 
\end{equation}

Where $K$ is the number of auxiliary classes. This idea is inspired by \citep{DBLP:journals/corr/MAXL}, and is useful to guard against the degenerate case where each primary task class has one and only one auxiliary class. 

The total reward is therefore constructed as:

\begin{equation}
Reward = \frac{-1}{B_R} \sum_{i=1}^{B_R} l_{primary}(f^{\theta}_{primary}(x_{(b),i}), y_{(b),i}) + \mathcal{H}(\pi_\theta(x)_{(b)})
\label{Rewardequation}
\end{equation}

Almost any RL algorithm/architecture can be used to define and update $\pi_{\theta}$. In practice, we used a customized Proximal Policy Optimization approach \citep{DBLP:journals/corr/SchulmanWDRK17}.    

\textbf{Label Hierarchy Optimization}
MAXL employs hierarchy-constrained label generation through the use of a Masked Soft-max \citep{DBLP:journals/corr/MAXL}. In other words, each class in the primary task will have some fixed number of subclasses in the auxiliary task. 

If $z$ are the logits generated from an input $(x_i,y_i)$ and $y_i$ represents the integer index of the primary label, then the following gives the probability for auxiliary label $k$:

\begin{equation}
    p_k = \frac{\exp(z_k)m_k}{\sum_{j=1}^{K} \exp(z_j)m_j}
\end{equation}

 \begin{equation}
 m_k = \begin{cases}
1 \text{ if } \psi \cdot y_i \leq k < \psi \cdot (y_i +1)\\
\text{else }0\\
\end{cases}    
 \end{equation}

Where $K$ is the number of auxiliary classes, $z$ represents raw logits, 
 and $m \in\{0,1\}^K$ acts as the mask. MAXL employs a hierarchy factor $\psi$ that dictates how many subclasses a primary task class will have in the auxiliary task. 

Focal loss is used after the Masked Soft-max to promote the use of the entire auxiliary label space. We have adopted this Masked Soft-max focal loss calculation approach for the Reinforcement Learning setup.

\textbf{Weight-Adjusted Approaches}
After initial successes with the RL task generation approach mentioned, we turned our attention to learning sample-level weight adjustment in both the Bi-Level Optimization and RL approaches. Our per-sample loss from Equation \ref{eq:sample_rl_loss} is adjusted slightly, so that the weighting is sample-specific:

\begin{equation}
\ell_{total}(x_i, y_i) = \ell_{primary}(f_{primary}^{\theta}(x_i), y_i) + \lambda_i\ell_{aux}(f_{aux}^{\theta}(x_i),\pi_{\phi}(x_i))
    \label{eq:weight_adjusted_rl_loss}
\end{equation}

\begin{figure}
    \centering
    \includegraphics[width=0.92\linewidth]{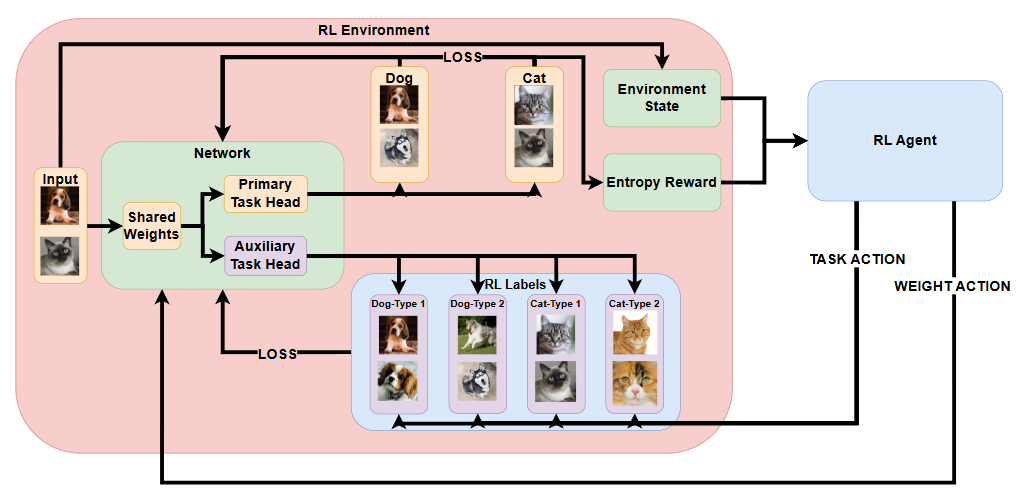}
    \caption{Weight-Aware Reinforcement Learning Auxiliary Task Setup}
    \label{fig:rl_weight-aware_diagram}
\end{figure}

For the Reinforcement Learning approach, we added another component to the action space, to enable the agent to select a weight for the given data point. The Weight-Adjusted Reinforcement Learning architecture is presented in Figure \ref{fig:rl_weight-aware_diagram}.

 For the Bi-Level Optimization approach, we took MAXL's open-source implementation and added a weight-output head to the labeling network such that it also outputs a weight parameter for a given sample.

\section{Methodology}

\textbf{RL Environment Implementation}
We implemented the main network environment (see Figure \ref{fig:rl_auxiliary_task}) using OpenAI's Gymnasium framework \citep{DBLP:journals/corr/BrockmanCPSSTZ16}. Our custom environment is configured to be agnostic of the primary task, type of main network, type of RL agent, system hardware, optimizer/scheduler, etc. These parameters can be set during instantiation, making our environment a robust platform to prototype Auxiliary Task training on different architectures and datasets rapidly.

\textbf{Agent Interaction}
Our environment is written to operate with an observation space that provides a single data point (image). The action it expects back is the RL agent's labeling of the last provided image. Therefore, on start-up, the environment samples a single batch from the training dataloader and provides a single data point (image) from the batch to the RL agent as the observation. The other data points in the batch are held for use as future observations. The RL agent then selects the action (label) for the observation and provides it to the environment. 

This continues back-and-forth until the environment has stored labels for each data point in the training batch (of size $B_T$). Once this happens, the main network is trained on that batch for both the primary and auxiliary tasks simultaneously using Equation \ref{eq:sample_rl_loss}. Then an evaluation batch (of size $B_R$) is sampled, and the reward is calculated as in Equation \ref{Rewardequation} and provided to the agent. A single episode consists of every single image in the training set being seen, labeled, and trained on once. An episode is equivalent to one epoch of training data.

\textbf{Training Modes}
\label{section:trainingmodes}
In addition to the standard OpenAI Gymnasium interface, our custom environment has additional features to support our unique training. This includes training evaluation functions, main network persisting, metric capturing, and so on.

Most importantly, our environment has been implemented to support two modes of training. They are:

\begin{itemize}
    \item \textbf{Train RL Agent Mode} - The agent is sampled as per their training algorithm. Throughout a training epoch, the weights of the RL agent and main network are both updated. The agent is pushed towards learning the policy that most improves the main network's performance on the primary task. At the end of the epoch, the main model is reverted to its starting state.
    \item \textbf{Train Main Network Mode} - The main model is trained with a static version of the auxiliary task. This is done by deterministically sampling the RL agent's policy for each image. Then the main model is trained on both tasks. As an optimization, the reward calculation is skipped as the agent is not being updated in this mode. At the end of the epoch, the main model is saved (and becomes the new "canonical model" for future RL Agent training).
\end{itemize}

\textbf{Weight Selection}
The current version of Stable-Baselines3's $MultiInputPolicy$ does not support combining discrete label actions ($MultiDiscrete$) and continuous scalar actions ($Box$) simultaneously \citep{stable-baselines3}. As such, when implementing the weight-aware version of the RL approach, we treat weight selection in the action space as a 21-class classification problem. Each class represented an evenly spaced value between 0 and 1, as in $\{0,0.05,0.1,0.15,...,0.95,1\}$. The selected unscaled weight $w_u$, is then used to get the true selected weight as:
\begin{equation}
    2^{10w_u-5}
\label{scaling_weight}
\end{equation}

This gives us a range of $[2^{-5},2^5]$ that the agent can select from. The weight-aware environment asks the agent to provide both a selection for the auxiliary task labeling and the weighting of the sample.

In the Weight-Aware MAXL implementation, the label network outputs a scalar value $[0,1]$ and is scaled the same as in Equation \ref{scaling_weight}.

\textbf{Datasets}
\label{section:datasets}
Cifar-10 and Cifar-100 have become ubiquitous benchmark datasets in the Auxiliary Learning space, both of which contain 60,000 images spread across 10 and 100 classes, respectively \citep{grégoire2023samplelevelweightingmultitasklearning,Joint_data_task, fan2018learning, DBLP:journals/corr/abs-1805-09501, DBLP:journals/corr/MAXL, DBLP:journals/corr/auxilearn, Krizhevsky2009LearningML}. 
Cifar-100 provides a 20 superclass hierarchy.  Each of the 100 classes in the dataset is mapped into one of the 20 Superclasses such that each superclass is composed of 5 of the 100 classes \citep{Krizhevsky2009LearningML}. Table \ref{Cifar100-20label} in the Appendix shows how the Superclasses are constructed. This is a very useful human-labeled auxiliary task benchmark, against which we can compare our proposed methodology. 
SVHN (Street View House Numbers) is also a prominent test dataset in the literature \citep{SVHN}. It is composed of 600,000 real-world images of address digits with 10 classes (0-9).
Our experiments will explore how the main network performance changes on these datasets as we introduce RL-created auxiliary tasks. For further training details, see Appendix B. For network details, see Appendix C.

\section{Results and Analysis}

\textbf{Reinforcement Learning vs MAXL}
To judge the performance benefits of our RL-based Auxiliary Task generation approach, we compare it to: (1) a network trained with a MAXL-generated Auxiliary Task, (2) a Human-Labeled Auxiliary Task (see Section \ref{section:datasets}), and (3) a network trained without an auxiliary task. We will test these approaches on the CIFAR-100 20-Superclass dataset. In this experiment, the auxiliary tasks being tested are equally weighted to the primary task.

\begin{figure}[htbp]
    \centering
    
    \begin{subfigure}[b]{0.48\linewidth} 
                \centering
        \includegraphics[width=\linewidth]{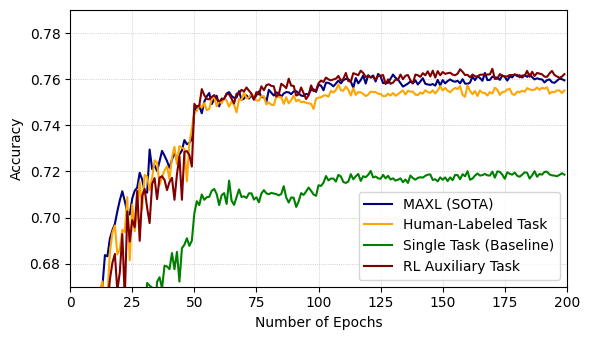}
        \caption{Test Accuracy over Epochs Plot}
        \label{fig:Wa-label-task}
    \end{subfigure}
    \hfill                   
    \begin{subfigure}[b]{0.48\linewidth}
        \centering
        \includegraphics[width=\linewidth]{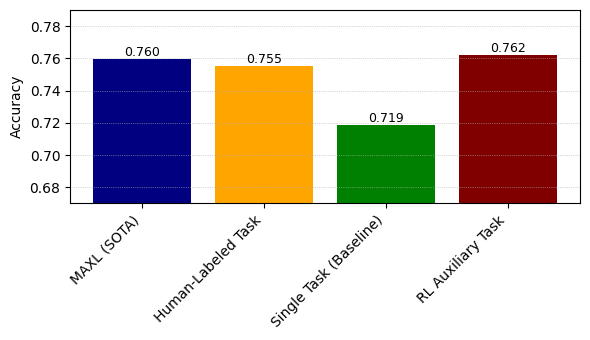}  
        \caption{Test Accuracy Bar Graph}
        \label{fig:rl_weight-aware}
    \end{subfigure}

    \caption{Cifar100 20-Superclass Performance of Auxiliary Task Methods}
    \label{fig:experiment_1_maxl_vs_rl}
\end{figure}

Figure \ref{fig:experiment_1_maxl_vs_rl} demonstrates that our RL-based auxiliary task generation approach (76.186\%) performs just as well (if not slightly better) than MAXL's Bi-Level Optimization approach (75.983\%). Both perform better than the human-labeled auxiliary task (75.525\%). All of these three are significantly better than the single-task baseline (71.864\%).

In Section \ref{sect:weight-aware experiments}, we will report results of the RL-based approach on additional datasets.

It should be noted that all trainings were redone for this paper. The values reported here are consistent with what the MAXL paper reported.

\textbf{Weight Adjustment Ablation}
Next we wanted to observe how tuning the task-level weight hyperparameter $\lambda$ affects the performance of the main network on the primary task. To do this, we retrained the RL Agent and the main network on the Cifar100 20-Superclass task with $\lambda=0.25,0.5,1,2,4$ and observed the results.

\begin{figure}[htbp]
    \centering
    
    \begin{subfigure}[b]{0.48\linewidth} 
                \centering
        \includegraphics[width=\linewidth]{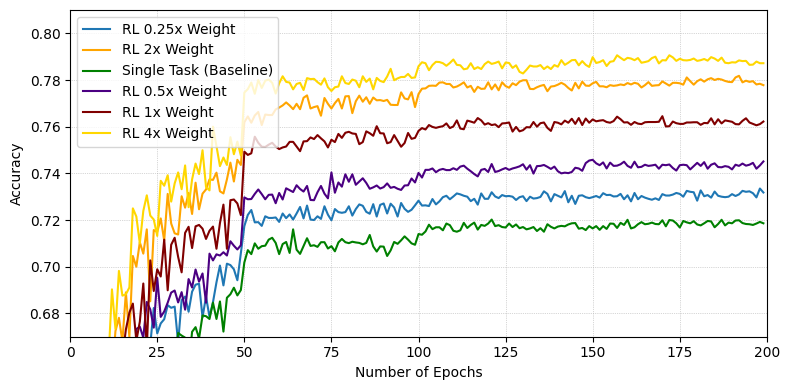}
        \caption{Test Accuracy over Epochs Plot}
        \label{fig:Wa-label-task}
    \end{subfigure}
    \hfill                   
    \begin{subfigure}[b]{0.48\linewidth}
        \centering
        \includegraphics[width=\linewidth]{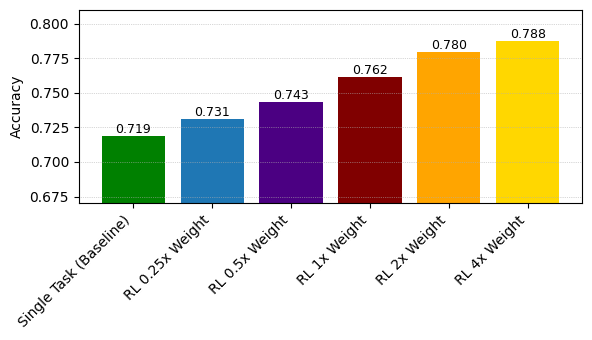}  
        \caption{Test Accuracy Bar Graph}
        \label{fig:rl_weight-aware}
    \end{subfigure}

    \caption{Cifar100 20-Superclass Auxiliary Task Weight Adjustment Ablation}
    \label{fig:rl_ablation_weight}
\end{figure}

Figure \ref{fig:rl_ablation_weight} shows that the performance of the main network on the primary task is sensitive to the weighting of the auxiliary task in the loss. Inspired by the clarity of the results from this ablation study, we turned our attention to methods that could learn sample-level weighting and auxiliary task labels jointly. We would like to be able to find an ideal weighting setup without an expensive hyperparameter search. This is what gave rise to our Weight-Aware MAXL and Weight-Aware RL approaches.

\begin{figure}[htbp]
    \centering
    
    \begin{subfigure}[b]{0.48\linewidth} 
                \centering
        \includegraphics[width=\linewidth]{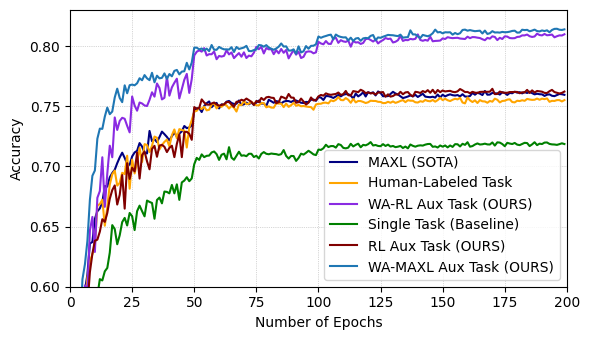}
        \caption{Test Accuracy over Epochs Plot}
        \label{fig:Wa-label-task}
    \end{subfigure}
    \hfill                   
    \begin{subfigure}[b]{0.48\linewidth}
        \centering
        \includegraphics[width=\linewidth]{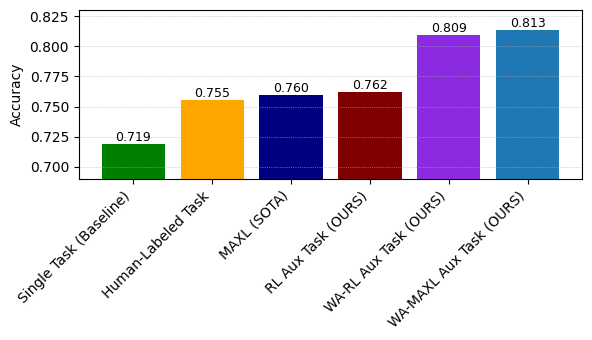}  
        \caption{Test Accuracy Bar Graph}
        \label{fig:rl_weight-aware}
    \end{subfigure}

    \caption{Cifar100 20-Superclass Weight-Aware Experiment Results}
    \label{fig:wa_tests_cifar100-20}
\end{figure}

\textbf{Weight-Aware Approach Experiment}
\label{sect:weight-aware experiments}
Now we wish to test the performance of our weight-aware approaches on the 20-Superclass CIFAR100 task. As is clear in Figure \ref{fig:wa_tests_cifar100-20}, our Weight-Aware MAXL (WA-MAXL) and Weight-Aware RL approach (WA-RL) provide significant performance improvement over the other methods. On the 20-Superclass Cifar100 task, WA-MAXL is the highest performer with 81.3\% accuracy, followed closely by WA-RL with 80.9\% accuracy. The non-weight-aware approaches hover in the 76\% range, while the single task network is just below 72\%

\begin{figure}[htbp]
    \centering
    
    \begin{subfigure}[b]{0.48\linewidth} 
                \centering
        \includegraphics[width=\linewidth]{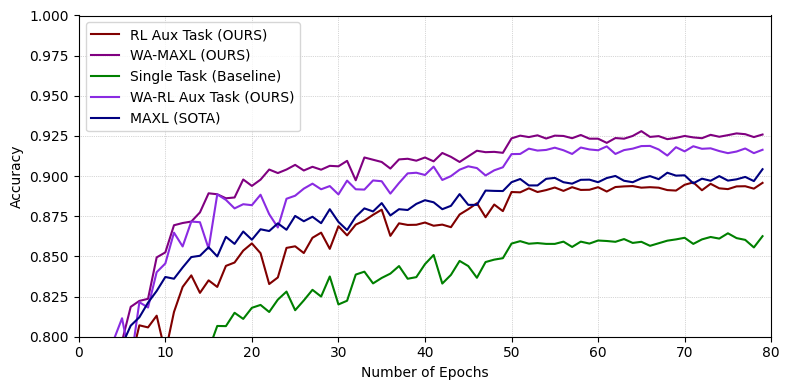}
        \caption{Test Accuracy over Epochs Plot}
        \label{fig:Wa-label-task}
    \end{subfigure}
    \hfill                   
    \begin{subfigure}[b]{0.48\linewidth}
        \centering
        \includegraphics[width=\linewidth]{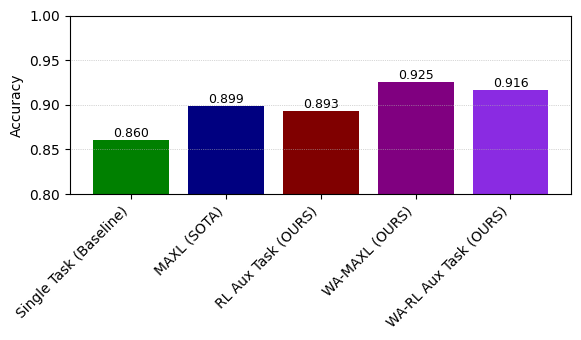}  
        \caption{Test Accuracy Bar Graph}
        \label{fig:rl_weight-aware}
    \end{subfigure}

    \caption{Cifar10 Weight-Aware Experiment Results}
    \label{fig:wa_tests_cifar10}
\end{figure}

We also performed experiments on the CIFAR10 dataset over 80 epochs. Figure \ref{fig:wa_tests_cifar10} shows that while weight-based approaches perform better, the performance gain is not quite as drastic. 

\begin{figure}[!htbp]
    \centering
    
    \begin{subfigure}[b]{0.48\linewidth} 
                \centering
        \includegraphics[width=\linewidth]{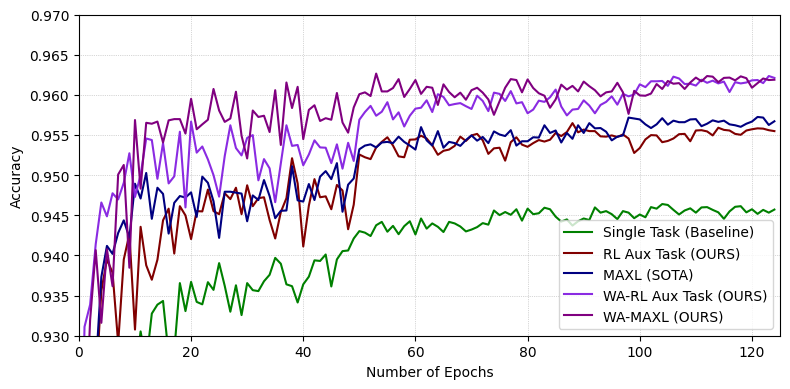}
        \caption{Test Accuracy over Epochs Plot}
        \label{fig:Wa-label-task_accuarcy_over_epochs}
    \end{subfigure}
    \hfill                   
    \begin{subfigure}[b]{0.48\linewidth}
        \centering
        \includegraphics[width=\linewidth]{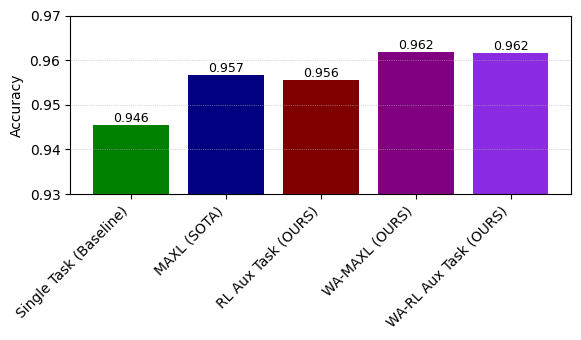}  
        \caption{Test Accuracy Bar Graph}
        \label{fig:rl_weight-aware_accuracy}
    \end{subfigure}

    \caption{SVHN Weight-Aware Experiment Results}
    \label{fig:wa_tests_svhn}
\end{figure}

Lastly, we test our approaches on the SVHN dataset over 125 epochs. As we can see in Figure \ref{fig:wa_tests_svhn}, once again, the weight-aware approaches perform best. Meanwhile, the regular RL and MAXL approaches still provide performance improvements over the baseline. There is a total improvement of about 1.5\%.

In summary, these experiments clearly demonstrate our framework's ability to improve a main network's performance on a target primary task. 

\section{Conclusion}
This work demonstrates that an RL agent can be used for auxiliary task generation without requiring the cost or complexity of Bi-level Optimization. We see that our RL-based approach outperforms a human-labeled auxiliary task and performs comparably to MAXL, one of the most prominent approaches in the space, across several benchmarks. We also demonstrate that sample-level auxiliary weights can be learned simultaneously with the auxiliary task selection. Both the Weight-Aware MAXL and RL approaches perform significantly better than their Weight-Unaware counterparts without requiring any additional hyperparameters or scaling cost. We hope to continue exploring the efficacy of RL in the Auxiliary Learning space in the future.

\label{headings}

\newpage
{
\small
\bibliographystyle{plainnat} 
\bibliography{references}  
}

\newpage
\appendix

\section{Appendix / supplemental material}

\begin{table}[ht]
\centering
\begin{tabular}{|l|p{10cm}|}
\hline
\textbf{Superclass} & \textbf{Classes} \\
\hline
Aquatic mammals & beaver, dolphin, otter, seal, whale \\
Fish & aquarium fish, flatfish, ray, shark, trout \\
Flowers & orchids, poppies, roses, sunflowers, tulips \\
Food containers & bottles, bowls, cans, cups, plates \\
Fruit and vegetables & apples, mushrooms, oranges, pears, sweet peppers \\
Household electrical devices & clock, computer keyboard, lamp, telephone, television \\
Household furniture & bed, chair, couch, table, wardrobe \\
Insects & bee, beetle, butterfly, caterpillar, cockroach \\
Large carnivores & bear, leopard, lion, tiger, wolf \\
Large man-made outdoor things & bridge, castle, house, road, skyscraper \\
Large natural outdoor scenes & cloud, forest, mountain, plain, sea \\
Large omnivores and herbivores & camel, cattle, chimpanzee, elephant, kangaroo \\
Medium-sized mammals & fox, porcupine, possum, raccoon, skunk \\
Non-insect invertebrates & crab, lobster, snail, spider, worm \\
People & baby, boy, girl, man, woman \\
Reptiles & crocodile, dinosaur, lizard, snake, turtle \\
Small mammals & hamster, mouse, rabbit, shrew, squirrel \\
Trees & maple, oak, palm, pine, willow \\
Vehicles 1 & bicycle, bus, motorcycle, pickup truck, train \\
Vehicles 2 & lawn-mower, rocket, streetcar, tank, tractor \\
\hline
\end{tabular}
\caption{Cifar-100 20 Superclass and Corresponding Single Classes}
\label{Cifar100-20label}
\end{table}

\subsection{MAXL Weight Ablation Study}

Similarly to how we conducted the ablation study to see the effect of changing the auxiliary task weight $\lambda$ on the RL setup, we conducted the same experiment for MAXL. 

\begin{figure}[htbp]
    \centering
    
    \begin{subfigure}[b]{0.48\linewidth} 
                \centering
        \includegraphics[width=\linewidth]{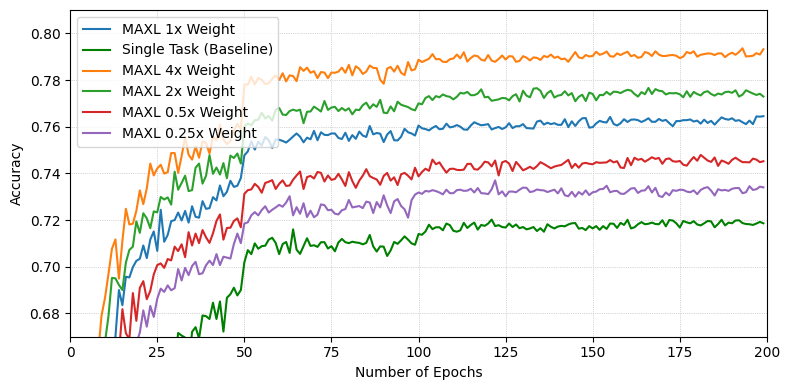}
        \caption{Test Accuracy over Epochs Plot}
        \label{fig:maxl-label-task}
    \end{subfigure}
    \hfill                   
    \begin{subfigure}[b]{0.48\linewidth}
        \centering
        \includegraphics[width=\linewidth]{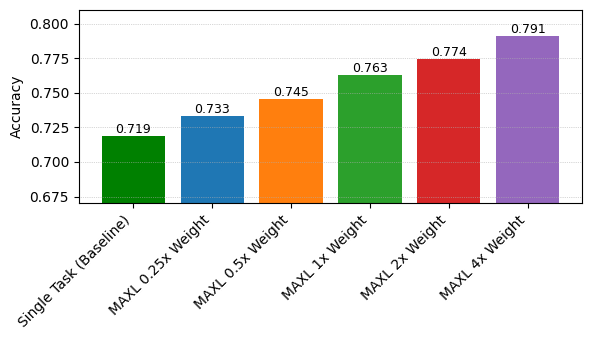}  
        \caption{Test Accuracy Bar Graph}
        \label{fig:maxl_weight-aware}
    \end{subfigure}

    \caption{Cifar100 20-Superclass Auxiliary Task Weight Adjustment Ablation on MAXL}
    \label{fig:maxl_ablation_weight}
\end{figure}

Figure \ref{fig:maxl_ablation_weight} shows a similar impact as in the RL case.

\subsection{Impact of Reset Ablation}
Instead of resetting the RL environment at the end of every epoch, we reset it at the end of every batch during the RL Agent training portion of the cycle. The main difference here is that over the course of one training epoch for the RL Agent, the main weights are constantly reset back to the canonical model.

\begin{figure}[htbp]
    \centering
    
    \begin{subfigure}[b]{0.48\linewidth} 
                \centering
        \includegraphics[width=\linewidth]{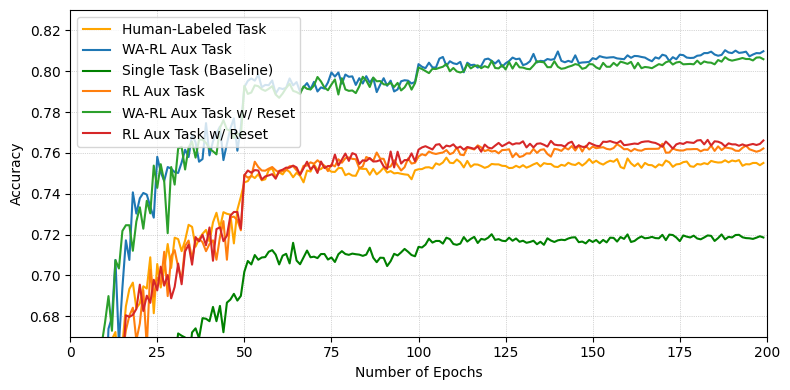}
        \caption{Test Accuracy over Epochs Plot}
        \label{fig:r_reset}
    \end{subfigure}
    \hfill                   
    \begin{subfigure}[b]{0.48\linewidth}
        \centering
        \includegraphics[width=\linewidth]{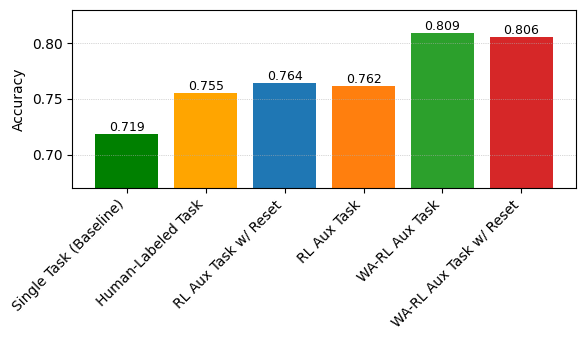}  
        \caption{Test Accuracy Bar Graph}
        \label{fig:rl_reset_bar}
    \end{subfigure}

    \caption{Cifar100 20-Superclass Auxiliary Task Environment Reset Ablation}
    \label{fig:rl_reset_ablation}
\end{figure}

The results in Figure \ref{fig:rl_reset_ablation} show that resetting the environment per batch does not yield different outcomes.

\subsection{Precision, Recall and F1 Score}

Accuracy is the main metric captured in the Auxiliary Learning literature. Most papers do not reference Precision, Recall, and F1 score in their analysis. For the sake of completeness, we have included some comparisons of these metrics. We include the Precision, Recall and F1 score for the RL approach on the 20-Superclass CIFAR100 problem. We also include comparisons of the baseline, the regular RL approach and the Weight-Aware RL approach on the SVHN dataset.

\begin{figure}[htbp]
    \centering
    
    \begin{subfigure}[b]{0.32\linewidth} 
                \centering
        \includegraphics[width=\linewidth]{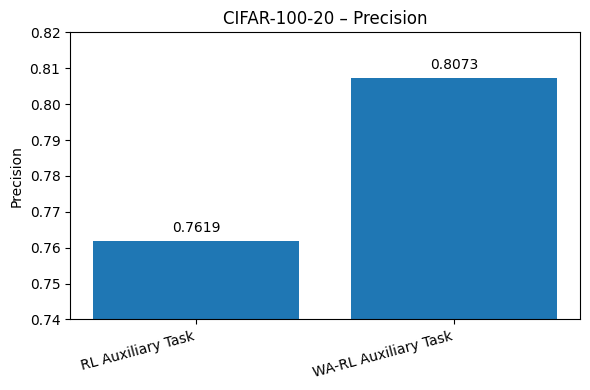}
        \caption{Precision Comparison}
    \end{subfigure}
    \hfill                   
    \begin{subfigure}[b]{0.32\linewidth}
        \centering
        \includegraphics[width=\linewidth]{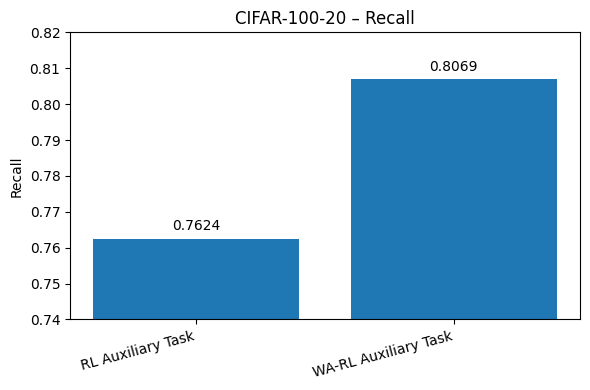}  
        \caption{Recall Comparison}
    \end{subfigure}
        \hfill                   
    \begin{subfigure}[b]{0.32\linewidth}
        \centering
        \includegraphics[width=\linewidth]{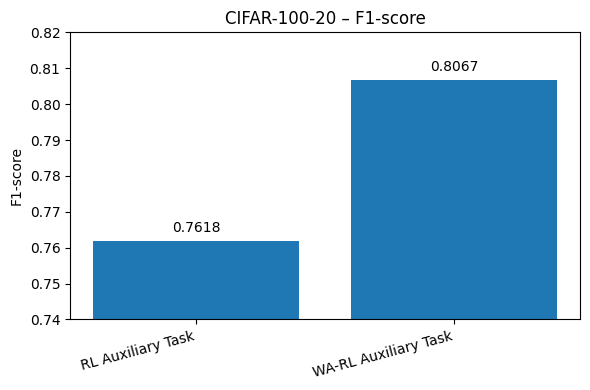}  
        \caption{F1 Score Comparison}
    \end{subfigure}

    \caption{20-Superclass CIFAR100 Precision, Recall and F1 Score Comparison}
    \label{fig:rl_reset_ablation}
\end{figure}

\begin{figure}[htbp]
    \centering
    
    \begin{subfigure}[b]{0.32\linewidth} 
                \centering
        \includegraphics[width=\linewidth]{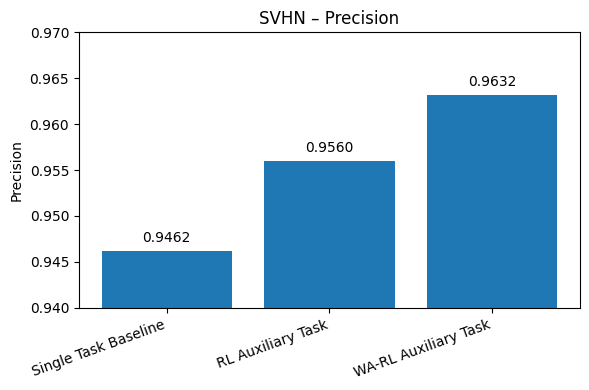}
        \caption{SVHN Precision Comparison}
    \end{subfigure}
    \hfill                   
    \begin{subfigure}[b]{0.32\linewidth}
        \centering
        \includegraphics[width=\linewidth]{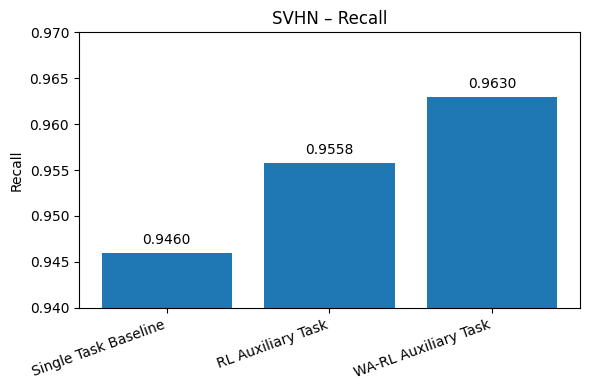}  
        \caption{SVHN Recall Comparison}
    \end{subfigure}
        \hfill                   
    \begin{subfigure}[b]{0.32\linewidth}
        \centering
        \includegraphics[width=\linewidth]{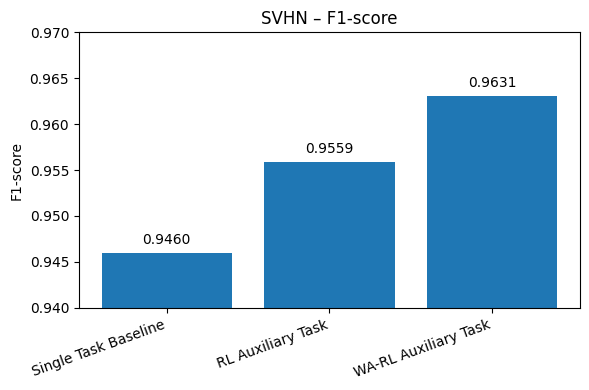}  
        \caption{SVHN F1 Score Comparison}
    \end{subfigure}

    \caption{SVHN Precision, Recall and F1 Score Comparison}
    \label{fig:rl_reset_ablation}
\end{figure}

\section{Training}
\begin{figure}
    \centering
    \includegraphics[width=0.7\linewidth]{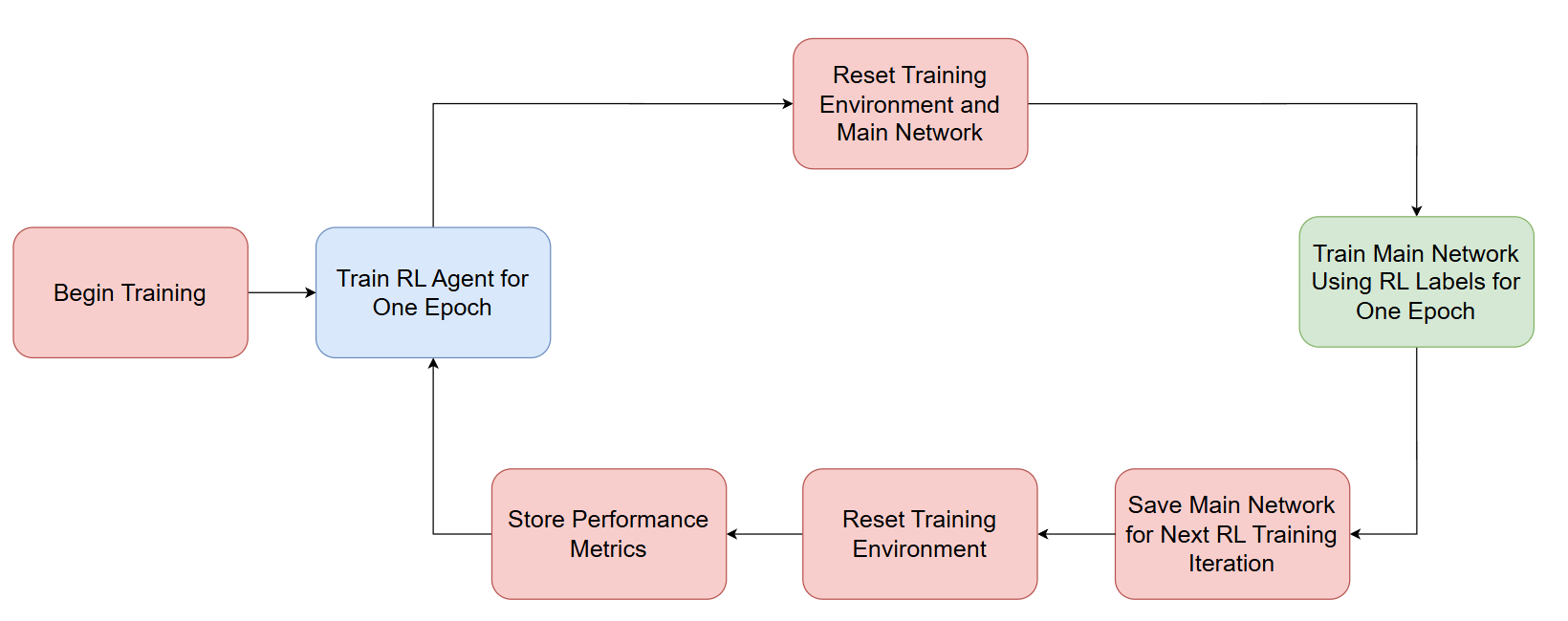}
    \caption{Alternating Training Approach}
    \label{fig:alternating training}
\end{figure}

The training procedure was standardized in all experiments, unless otherwise noted. The training strategy involved alternating the training between one epoch of RL Agent training followed by one epoch of main network training. The two supported training modes mentioned in Section \ref{section:trainingmodes} facilitated this alternating training strategy. The environment was reset (dataloader was reshuffled, canonical main network was loaded, global state cleared, etc.) between batches. Figure \ref{fig:alternating training} visualizes this workflow.

SGD with a learning rate scheduler was used as the optimizer to train the main network. The standard transformations (cropping, normalization, etc.) for each dataset were applied. Early stopping was used to obtain performance metrics. The following are the default hyperparameters that were used in all training, unless otherwise stated: Training Batch Size$(B_T)$=100, Evaluation Batch Size$(B_R)$=256,Feature Extraction Dimensions=256, Number of Epochs=200, Primary Learning Rate=0.01, PPO Learning Rate=0.0003, PPO Entropy Coefficient=0.01, Scheduler Step Epochs=50, Scheduler Gamma=0.5, Auxiliary Task Weight$(\lambda)$=1, Hierarchy Factor$(\psi)$=5.

The training was conducted on several 2080Ti GPUs simultaneously. Each experiment took between 12-36 hours to run. About 800-900 GPU hours were required to conduct this research.

\section{Networks}

\subsection{Main Network}

As it is widely used in the Auxiliary Learning literature, we focus our experiments on a custom implementation of the VGG16 architecture \citep{VGG16, DBLP:journals/corr/auxilearn, DBLP:journals/corr/MAXL, DBLP:journals/corr/grad_norm, shamsian2023auxiliarylearningasymmetricbargaining}. The primary difference is that there are two classifier heads (for the primary and auxiliary tasks). Each head is comprised of two linear layers. The first projects the features from the final convolutional layer to a 512-dimensional representation, and the second maps this representation to the respective task’s output dimension.

\subsection{Agent Network}

We used a customized Proximal Policy Optimization (PPO) Algorithm to learn the auxiliary task provided to the main network \citep{DBLP:journals/corr/SchulmanWDRK17}.
PPO is an on-policy method that was created by OpenAI. It attempts to prevent large policy changes by optimizing:
\begin{equation}
L^{\mathrm{CLIP}}(\theta) = \mathbb{E}_t \!\Big[ \min\bigl(r_t(\theta) A_t,\, \operatorname{clip}\bigl(r_t(\theta), 1-\epsilon, 1+\epsilon\bigr)\,A_t \bigr)\Big],
\end{equation}
where 
\begin{equation}
r_t(\theta) = \frac{\pi_{\theta}(a_t \mid s_t)}{\pi_{\theta_{\mathrm{old}}}(a_t \mid s_t)}
\end{equation}
and \(A_t\) denotes the advantage.

For ease of integration with our Gymnasium environment, we implemented the policy using the open source Stable-Baselines3 framework \citep{stable-baselines3}. Our implementation mirrors the main network as it uses the same architecture (VGG16) as the backbone.

Concretely, our implementation is composed of 3 custom components:

\begin{itemize}
    \item Feature Extractor Network - VGG16 architecture that takes in an image and converts it to a 256-dimensional representation.
    \item Action Network - Linear mapping layer that takes in 256-dimensional representation from the Feature Extractor Network and outputs a vector of the size of the auxiliary task dimension. The output represents the agent's auxiliary label selection for the given image.
    \item Value Network - Linear mapping layer that takes in 256-dimensional representation from the Feature Extractor Network and outputs a single value, representing the value of the observation.
\end{itemize}

\end{document}